\documentclass[a4paper]{article}

\usepackage{INTERSPEECH2016}

\usepackage{graphicx}
\usepackage{amssymb,amsmath,bm}
\usepackage{textcomp}
\usepackage{color}

\usepackage{amssymb}
\usepackage{amsmath}
\DeclareMathOperator*{\argmax}{argmax}
\newcommand{\PJ}[1]{{\color{black}#1}}

\ninept

\title{Performance Improvements of Probabilistic Transcript-adapted ASR with Recurrent Neural Network and Language-Specific Constraints}

\makeatletter
\def\name#1{\gdef\@name{#1\\}}
\makeatother \name{{\em Xiang Kong$^1$, Preethi Jyothi$^2$, Mark Hasegawa-Johnson $^2$}}

\address{$^1$ Department of Computer Science, University of Illinois at Urbana-Champaign, IL, USA \\
  $^2$Beckman Institute, University of Illinois at Urbana-Champaign, IL, USA \\
  {\small \tt \{xkong12, pjyothi, jhasegaw\}@illinois.edu}
}

\begin{document}

  \maketitle
  \begin{abstract}
    Mismatched transcriptions have been proposed as a mean to acquire probabilistic transcriptions from non-native speakers of a language. Prior work has demonstrated the value of these transcriptions by successfully adapting cross-lingual ASR systems for different target languages. In this work, we describe two techniques to refine these probabilistic
    transcriptions: a noisy-channel model of non-native phone misperception
    is trained using a recurrent neural network, and decoded using
    minimally-resourced language-dependent pronunciation constraints.
    Both innovations improve quality of the transcript, and both innovations
    reduce phone error rate of a trained ASR, by 7\% and 9\% respectively.
  \end{abstract}

\section{Introduction}

In building standard automatic speech recognition (ASR) systems, the first step is to collect sufficient amounts of transcribed speech data. Crowdsourcing \cite{Scott-Novotney} \cite{Maxine} has been regarded as a useful resource to speed up this step, by acquiring transcripts from a large number of crowd workers. However, normal crowdsourcing methods fail for under-resourced languages since they require transcribers to be native speakers of the target language, and it is hard to find native speakers of under-resourced languages online. As a matter of fact, according to the survey \cite{Laurent-Besacier}, most languages are under-resourced which makes crowdsourcing only applicable to a small fraction of the languages in the world.
	
    Mismatched crowdsourcing was introduced as a potential way to solve the limitation of crowdsourcing. In mismatched crowdsourcing, workers transcribe a language they do not know at all. Instead of using meaningful words in the target language, they just use nonsense syllables in their own orthography (here: English letters) to transcribe speech data \cite{Pr}. These so-called mismatched transcripts are decoded using a noisy channel model \cite{Pr} \cite{PreInter}  in order to derive a probabilistic transcription (PT) \cite{chunxi}, defined as a probability mass function (pmf) over phoneme sequences in the utterance language.
	
    Although PTs are not as accurate as native transcriptions, it has been shown that ASR adapted using PTs is more accurate than cross-lingual or semi-supervised ASR \cite{chunxi}. 
	
    This paper describes two methods that further improve the error rate of ASR trained using PTs. First, we have previously shown \cite{xiang} that language specific constraints reduce the error rate of a PT, but \cite{xiang} only focuses on the text level; this paper demonstrates that the resulting PTs also yield ASR with reduced error. Second, this paper re-imagines probabilistic transcription as a kind of machine translation problem, in which mismatched transcripts are translated to phone sequences. Building on this analogy, sequence-to-sequence translation models based on recurrent neural networks (RNNs) were trained to generate PTs, from which ASR was trained. 
	
    Section 2 briefly describes probabilistic transcriptions, the language-specific constraint method, and the RNN-based PT decoding scheme. Section 3 describes data preparation, along with the experimental details and results, and our main conclusions are listed in Section 4. 

  \section{Improving and Constraining the Channel for Mismatched Crowdsourcing}

  A probabilistic transcription (PT) is a distribution over utterance-language phone sequences.
  When a PT is computed from mismatched crowdsourcing, it can be written as 
  $\Pr(\pi|T)$, where $T$ is a set of mismatched transcripts in the orthography
  of the annotation language (e.g., English), and $\pi$ is a
  phone sequence in the orthography of the utterance language
  (e.g., Greek).  In order to make the computation of $\Pr(\pi|T)$ tractable,
  the nonsense texts produced by multiple transcribers are first pruned
  and merged in order to create a distribution over nonsense texts~\cite{Pr}.
  Using $\lambda$ to denote this annotation-language nonsense transcript
  when viewed as a random variable, the computations can be summarized as:
      \begin{equation}
      \begin{split}
        \Pr(\pi |T)&=\sum_\lambda \Pr(\pi, \lambda | T)\Pr(\lambda |T)\\
        &\approx max_\lambda \Pr(\pi |\lambda)\Pr(\lambda |T)\\
        &=max_\lambda \left (  \frac{\Pr(\lambda |\pi)}{\Pr(\lambda)}\Pr(\pi)\right )\Pr(\lambda |T)
        \label{eq1}
      \end{split}
      \end{equation}
      $\Pr(\lambda|\mathbf{\pi})$ is represented by a FST which maps phones to letters \cite{Pr}. $\Pr(\lambda)$ is a prior over transcriber-language letter sequences, $\Pr(\lambda |T)$ is a distribution to merge the transcripts in T over "representative transcripts" denoted by $\lambda$， and $\Pr(\pi)$ is a bigram phone language model trained using phone sequences synthesized in the utterance language by applying a character-based grapheme-to-phoneme converter to some utterance-language texts downloaded from Wikipedia.
      As for the decoding process, the phone sequence which maximizes $\Pr(\mathbf{\pi}|T)$ will be chosen according to the following equation:
      \begin{equation}
        \pi^{*}= \operatorname*{arg\,max}_\pi\Pr(\pi|T)
        \label{eq2}
      \end{equation}
     \subsection{Target-Language Post-Processing}

        A PT is a probability distribution, in a form that can be  represented by a weighted finite state transducer (WFST). Language specific constraints have been applied to reduce the entropy of PTs. Although these constraints are mined from Internet text, the results in \cite{xiang} showed that they can reduce the error rate of PTs.
		
        Since we have assumed that a pronunciation dictionary is unavailable for the target language, a list of orthographic words in the target language without known pronunciation is used instead. Since available text data in the target language is sparse, we assume every word in this list is equally likely. This language model constraint can be represented as an FST, denoted as \textbf{LM}. Although there may not exist a suitable pronunciation model, downloading and appropriately reconstructing the Wikipedia page titled "XXX Alphabet"(where XXX is the test language, for example, Greek in this case), is a possible method to estimate the set of all possible grapheme-to-phoneme (G2P) mappings for the target language. Then for any given grapheme sequence in the test language, the resulting G2P produces all phone sequences (with equal likelihood) that are attested on Wikipedia as possible pronunciations. This kind of unweighted G2P is unlikely to be useful for automatic speech recognition, but is better than having no pronunciation at all: it imposes  a loose upper bound on the collection of possible phone sequences that might correspond to any given orthographic sequence. Such G2Ps have been constructed for seventy languages, and are available at \cite{mark}. In this paper, we denote the resulting FST mapping graphemes to phonemes as \textbf{G2P}.
		
        We regard the FSTs \textbf{G2P} and \textbf{LM} described above as the available target language information. It is possible to reduce PERs of PTs by only applying this information as constraints on the PTs. The information in these transducers is applied to the \textbf{PT} by composing the FSTs, creating the modified probabilistic transcription \def\comp{\ensuremath\mathop{\scalebox{.6}{$\circ$}}} $\widehat{\textbf{PT}}=\textbf{G2P}^{-1}\circ\textbf{LM}\circ\textbf{G2P}\circ\textbf{PT}$
		
		Furthermore, since the weights on each edge in the G2P and LM are regarded as not useful, it is possible to discount them using zero-exponentiation, defined such that $x^0\triangleq 1$ if and only if $x\neq 1$ but $0^{0}\triangleq 0$. Using this notation, the edge weights in \textbf{G2P} are a model of Pr$(W|\pi)^{0}$ and those in \textbf{LM} are a model of  Pr$(W)^{0}$, therefore the edge weights in $\widehat{\textbf{PT}}$ are a model of
		\begin{equation}
          \widehat{PT}(\pi|T)=\sum\limits_{W}\Pr(\pi|W)^0\Pr(W)^0\Pr(W|\pi)\Pr(\pi|T)
		  \label{eq3}
        \end{equation}
        where \textbf{$W$} is the possible word sequence. Eq. (3) shows that \textbf{PT} with post processing is a combination of a language-dependent G2P (representing the set of grapheme-to-phoneme transductions that are attested in the target language) and a language-dependent language model LM.  As an intermediate step, it is possible to limit PT by simply pruning away phonemes that don't exist in the test language.  Phoneme pruning can be computed by the FST composition \def\comp{\ensuremath\mathop{\scalebox{.6}{$\circ$}}} $\widehat{\textbf{PT}}=\textbf{G2P}^{-1}\circ\textbf{G2P}\circ\textbf{PT}$, which computes:
		\begin{equation}
          \widehat{PT}(\pi|T)=\sum\limits_{W}\Pr(\pi|W)^0\Pr(W|\pi)^0\Pr(\pi|T)
		  \label{eq4}
        \end{equation}
	\subsection {RNN-based Probabilistic Transcriptions}
	    \PJ{Recurrent neural networks (RNNs) have been successfully applied to a number of sequence prediction tasks. A popular RNN model   called RNN Encoder-Decoder was introduced to learn mappings between variable-length sequences which was successfully evaluated on a machine translation task~\cite{Cho}. We observe that the problem of building probabilistic transcriptions could be viewed as a simpler variant of the translation problem in which mismatched transcripts are translated into phone sequences.}

\PJ{The RNN Encoder-Decoder architecture consists of a pair of RNNs -- an encoder and a decoder RNN. The encoder RNN processes an entire input sequence sequentially and produces a fixed-length representation after reading the last symbol of the input sequence. The decoder RNN then decodes the given fixed-length vector into a target-language phone sequence. During the training process, the probability of a source sentence given a target sentence is maximized by jointly training the two RNNs.}
 
Formally, we denote the source symbol sequence, as $X=\{ x_1,x_2,...,x_T\}$, and the target sequence as $Y=\{ y_1,y_2,...,y_{\hat{T}}\}$ (the lengths of X and Y could be different). Following notation introduced in~\cite{Cho}, the encoder reads each input symbol into an RNN in which the hidden state $h_{<t>}$ is updated as:
		\begin{equation}
          h_{<t>}=f(h_{<t-1>},x_{t})
		  \label{eq5}
        \end{equation}
		where function f is a non-linear activation function.
		
		\PJ{The encoder produces a fixed-length vector (denoted by $c$) as output after reading the entire input sequence; the last symbol of the input sequence is marked by a special end-of-sequence symbol e.g. "$\textless/s\textgreater$". $c$ is an encoded representation of the input sequence that is obtained from the hidden state of the encoder RNN at time step $T$. Conditioned on this summary vector $c$, the hidden states in the decoder RNN are updated by:} 
		\begin{equation}
          h_{<t>}=f(h_{<t-1>},y_{t-1},c)
		  \label{eq6}
        \end{equation}
\noindent where $h_{<t>}$ is also conditioned on the previous symbol in the target sequence, $y_{t-1}$. The decoder RNN produces the target sequence according to the following probability distribution (conditioned on both $c$ and $y_{t-1}$): 
		\begin{equation}
          \Pr(y_t|y_{t-1},...,y_1,c)=g(h_{<t>},y_{t-1},c)
		  \label{eq7}
        \end{equation}
\noindent Here, $g$ is the softmax activation function. The encoder and decoder RNNs are trained by choosing network parameters in order to maximize the log-probability of the output sequence given the input sequence. i.e.
		\begin{equation}
                  \theta = \argmax \frac{1}{N} \sum_{n=1}^{N}\log(\rm{\Pr}_{_{\theta}}(Y_n|X_n))
		  \label{eq8}
        \end{equation}
where $N$ is the number of training samples and $\theta$ is the set of parameters in the model. 

\PJ{In this paper, the encoder and decoder RNNs are implemented using Long Short-Term Memory (LSTM) networks, which can learn relatively longer temporal dependencies than simple RNNs. More experiment details are described in section 3.3.}

        PTs are computed by decoding mismatched transcripts using a WFST model of non-native speech perception. The non-native perceptual model can also be viewed as a sequence-to-sequence translation of non-native phones into letters in the native orthography.  Since this process is analogous to translation, it is possible to train and test algorithms that have been effective in machine translation, such as the encoder-decoder RNN architecture. During training, the source sequence is a set of annotation-language orthographic transcripts (acquired from crowd workers), while the target sequence is an utterance-language phone sequence (acquired from a native speaker of the utterance language). In our target application, it may not be possible to recruit native speakers of the target language, therefore the RNN is trained using transcripts of speech in other languages. For example, to train an RNN that translates English mismatched transcription into Greek phones, we use training data that includes English mismatched transcription of six other languages: Arabic, Cantonese, Dutch, Hungarian, Mandarin, Swahili, and Urdu. Vocabulary for input is English letters (lower case and capital case), and output is International Phonetic Alphabet (IPA). 
		
        After training this model by equation (8), we apply it to translate the input sequence $X$ to the target language phone sequence $Y$. Instead of using phone sequences directly from this model, we added a bi-gram language model mapping the target language into the translation process, which could (1) remove phonemes which never occur in the target language, and (2) improve accuracy:
		\begin{equation}
          Y^* = \argmax_{Y} \sum_{t=1}^{T}\left[ \log \rm{\Pr}_\theta(y_t|y_{t-1},c) + \log \rm{\Pr}_{_{LM}}(y_t|y_{t-1})\right]
		  \label{eq9}
        \end{equation}
    where T is the length of the translated sequence $Y$ and LM is the bigram language model of the target language.    
	\section{Experiments and Results}

        Speech data were extracted for eight languages from SBS audio podcasts \cite{SBS}: Arabic (ARB), Cantonese (YUE), Dutch (NLD), Greek (ELL), Hungarian (HUN), Mandarin (CMN), Swahili (SWH) and Urdu (URD). Native transcriptions were acquired for 40-60 minutes of speech in each of these languages. Cantonese transcriptions were obtained from I2R, Singapore as part of a collaborative research project, Greek transcriptions were collected from native speakers of Greek on the crowdsourcing platform UpWork \cite{upwork} and speech in the remaining six languages were transcribed by paid student volunteers at the University of Illinois, Urbana-Champaign who were native speakers of these languages. All the native transcriptions were converted into phonemic sequences using a universal phone
        set, manually constructed as a subset of the IPA.
        The eight speech datasets were split into roughly 40 minutes of training data and 10 minutes each for the development and evaluation sets.
	  
	  Mismatched transcriptions were acquired and decoded using published methods \cite{chunxi}. Ten different crowd workers on Amazon’s Mechanical Turk \cite{amazon} \cite{Ellie} were asked to provide a sequence of English nonsense syllables for each audio clip.
	  
	  \textbf{Cross-lingual baseline ASR systems}: For all eight languages, we built standard GMM-HMM based ASR systems trained on speech from all languages other than the language being recognized. These ASR systems were implemented using the Kaldi toolkit \cite{dan} and are henceforth referred to as \textbf{CL}.
	  
	  \textbf{ASR adaptation}: The cross-lingual baseline ASR systems (trained on data other than the target language) were adapted with probabilistic transcriptions for the target language using maximum a posteriori (MAP) estimation. This adaptation process is detailed further in \cite{chunxi}.
	  
    \subsection{Using Language-Specific Pronunciation Constraints}

      We demonstrate the utility of language-specific G2P mappings using Greek as our target language. Since there is no standard pronunciation dictionary available for Greek, we resort to two simple techniques to derive pronunciation constraints for Greek, described first in \cite{xiang}. 
	  
      First, a set of G2P mappings were compiled for Greek based on the description of its orthography \cite{mark}. These rule sets can be fairly easily generated for a range of different languages, as shown in \cite{mark}. Second, a Greek pronunciation dictionary was constructed using data available from the {\em Translation as a Service (Taas)} project \cite{tass}, which has about 200,000 Greek words with corresponding pronunciations available for some words. These pronunciations use an ASCII-based alphabet which we further map to IPA using a deterministic phone mapping. 

      Specifically, we defined three kinds of constraints \cite{xiang}. 
      \begin{itemize}
        \item We first used the inverse of the Greek G2P rules to map the phone-based transcriptions into Greek word sequences and then map them back to phone sequences for evaluation (computing the (PT) transducer of Eq. (4)). This simple constraint forces the probabilistic transcriptions to be matched with valid Greek words. This is referred to as ``\textbf{G2P}".
        \item For the sake of comparison, we also compute phone error rates using the Greek dictionary (described above) in conjunction with the G2P rules (computing the (PT) transducer of Eq. (3)). Greek words appearing in the dictionary are mapped to their corresponding pronunciations and any remaining out-of-vocabulary words are mapped to phones using the Greek G2P rules. This constraint is referred to as ``\textbf{G2P} + \textbf{dict}".
        \item Finally, we also compute the (PT) transducer after imposing word constraints using a bigram word language model (i.e., obtained by replacing $\Pr(W)^{0}$ in Eq. (3) with $\Pr(W)$ derived from a bigram model). Similar to the phone bigram model, we use the Greek Wikipedia text and the CMU CLMTK toolkit to train a bigram word-level language model. We refer to this constraint as ``\textbf{WLM}".
       
      \end{itemize}

    Bigram phone language models for Greek were implemented to represent Pr($\pi$) in Equation (1). In order to train a phone language model, the Greek word sequences were first mapped to phone sequences using the Greek pronunciation dictionary described earlier, along with applying the Greek G2P rules for any out-of-vocabulary words.
	
    Table~\ref{tab:1} shows the ASR adaptation results on the Greek development and evaluation sets after using each of the above-mentioned pronunciation constraints.  We observe a consistent reduction in phone error rates after incorporating the G2P constraints. Using PT adaptation with the cross-lingual baseline system significantly improves PER performance with a relative PER reduction of 9\%.  Adaption with PTs subjected to the G2P constraints further improves ASR performance, with the WLM constraints providing the largest reduction of 10\% PER on the evaluation set.
    
      \begin{table}[th]
        \caption{\label{tab:1} {\it PERs on the evaluation and development sets of Greek using various language-specific constraints.  CL=cross-lingual baseline.  PT=PT adaptation.  G2P, dict, WLM: defined in the text.}}
        \vspace{2mm}
        \centerline{
          \begin{tabular}{| c | c | c |}
            \hline
            \multicolumn{1}{|c|}{System} & 
            \multicolumn{1}{|c|}{PER\% (dev)} & 
            \multicolumn{1}{c|}{PER\% (eval)} \\
            \hline \hline\
              CL & 71.39 &       69.68~~~ \\
              CL + PT & 64.72 &         63.69~~~ \\
              CL + PT + G2P & 62.07 & 61.45~~~ \\
              CL + PT +G2P + dict & 60.24 &        60.11~~~ \\
              CL + PT + WLM & 58.96 &        57.23~~~ \\
            \hline
          \end{tabular}
        }
      \end{table}

      \subsection {Encoder-decoder RNN-based PTs}

          \begin{table*}[th]
        \caption{\label{tab:2} {\it Phone error rates of the best path through the PTs based on RNN sequence-to-sequence models, compared to baseline using an FST sequence-to-sequence model (from \cite{chunxi}).  Performance on development set in parentheses.}}
        \vspace{2mm}
        \centerline{
          \begin{tabular}{| c | c | c | c |}
            \hline
            Language (ISO 639-3) &
            FST Translation &
            RNN Translation &
            \% Rel. Reduction \\
            \hline \hline\
              ARB &	66.2 (65.8) &	60.9 (60.3) 	&8.0 (8.3)\\
              YUE &	67.8 (66.4)	&62.4 (62.7)	&8.0 (5.6)\\
              NLD &	70.9 (68.9)	&67.9 (64.8)	&4.2 (6.0)\\
              HUN &	63.5 (63.7)	&60.2 (59.6)	&5.2 (6.4)\\
              CMN &	69.6 (70.9)	&67.2 (66.8)	&3.4 (5.8)\\
              SWH &	50.3 (47.6)	&46.8 (43.6)	&7.0 (8.4)\\
              URD &	70.5 (67.2)	&66.9 (64.6)	&5.1 (3.9)\\
            \hline
          \end{tabular}
        }
      \end{table*}
      
      \begin{table*}[th]
        \caption{\label{tab:3} {\it Phone error rates of ASR adapted to PTs.  PTs constructed using either FST-based (numbers from \cite{chunxi}) or RNN-based translation models.  Dev set in parentheses.}}
        \vspace{2mm}
        \centerline{
          \begin{tabular}{| c | c | c | c | c |}
            \hline
            \multicolumn{1}{|c|}{Language code} & 
            \multicolumn{1}{|c|}{Cross-lingual (CL)} & 
            \multicolumn{1}{|c|}{CL + PT-FST adaptation} & 
            \multicolumn{1}{p{3cm}|}{CL +  RNN –based PT adaptation} & 
            \multicolumn{1}{p{3cm}|}{\% Rel. redn over CL+PT-FST} \\
            \hline \hline\
             YUE &	68.40 (68.35) &	57.20 (56.57) &	55.62 (56.01) &	2.8 (1.0)\\
             HUN &	68.62 (66.90) &	56.98 (57.26) &	53.73 (54.32) &	5.8 (5.1)\\
             CMN &	71.30 (68.66) &	58.21 (57.85) &	55.64 (55.90) &	4.4 (3.4)\\
             SWH &	63.04 (64.73) &	44.31 (48.88) &	41.21 (44.66) &	7.0 (4.2)\\
            \hline
          \end{tabular}
        }
      \end{table*}

      For the encoder-decoder RNN-based PTs, the experimental setup was designed to be the same as in \cite{chunxi} in order to enable a fair comparison. Seven languages other than Greek were used to train the RNN, which was then tested using Greek.
      
	Every input and output symbol was encoded using a one-hot representation. The input symbols are English letters and output symbols are phones in IPA. We also introduced two special symbols marking the start of a sentence, "$\textless s \textgreater$" and the end of a sentence, "$\textless /s \textgreater$". There was no unknown word symbol since both symbol sets were derived from a fixed vocabulary. The input sequence was reversed in order to improve performance (as suggested in \cite{Sut}). We used two deep LSTMs  as our basic model for the encoder and the decoder, respectively and each of them have two hidden layers. Each hidden layer comprised 100 hidden units. A softmax activation function is used at the network output to generate probability distributions across the output phones. The LSTM parameters were initialized using a uniform distribution between -0.1 and 0.1. Stochastic gradient descent with a fixed learning rate of 0.4 was used to train our LSTM network, with a convergence threshold of e=$10^{-7}$. For each iteration, 128 mismatched transcripts and their aligned native transcripts in the training set were selected for training. After 8 epochs, the learning rate was adjusted to 0.2 for more precise updating. These hyper-parameter settings were determined by fine-tuning on the development sets. After training this model, we used a small amount of target language data (10 minutes) to adapt this model further. The LSTM architecture was implemented using Theano libraries [18] and run on GPUs.
	
	Training data include 11520 mismatched transcripts and their corresponding native phonetic transcriptions in languages other than the target language. The development data include 480 mismatched transcriptions and their native phonetic transcriptions in target languages.
	
	For each input symbol in the mismatched transcriptions, the RNN models compute a probability distribution over the output phones. This output distribution is implemented using a finite state transducer and composed with a phone bigram language model for each language. The bigram language models are constructed using text documents from Wikipedia mined for each of these languages. The word sequences were converted into phone sequences using G2P/pronunciation models before training bigram phone language models (using the CMU CLMTK toolkit \cite{CMU}). More details on building these phone language models are specified in \cite{chunxi}. 
	
	Table~\ref{tab:2} shows the 1-best probabilistic phone transcription error rates using PTs derived from the RNN-based models. The numbers in parentheses refer to the 1-best error rates computed using PTs derived using maximum-likelihood alignments between letter sequences and phone sequences trained using the EM algorithm (as defined in \cite{chunxi}); we will refer to them as PT-ML. We observe that the RNN-based PTs consistently outperform PT-MLs across all seven languages. The largest PT error reductions were observed for Swahili and Arabic. 
	
	Table~\ref{tab:3} demonstrates that the improvement in quality of RNN-based PTs translates to improvements in ASR performance. We see performance improvements in ASR systems built for four different languages, with the largest improvement observed for Swahili.
        
\section {Conclusion}
    This article describes two methods to reduce phone error rates of probabilistic transcriptions (PTs) built from mismatched crowdsourcing by adding language specific constrains, (i.e. a rudimentary rule-based G2P and a list of orthographic word forms mined from the internet) and applying an RNN Encoder-Decoder architecture. ASR built on multilingual transcriptions is adapted by these two optimized PTs. Compared with simple PT-adapted ASR, the post processing method trains an ASR with up to 8.9\% relative PER reduction, while the RNN Encoder-Decoder based PT (without any language specific constraints) provides up to 7.0\% relative PER reduction.

  \section{Acknowledgements}
  
   This work was funded in part by a grant from the DARPA LORELEI program.

  \bibliographystyle{ieeetr}
  \newpage
  \bibliography{mybib_mh_2016mar28}
   

\end{document}